\pdfoutput=1
\documentclass[11pt]{article}

\usepackage[preprint]{acl}

\usepackage[most]{tcolorbox}
\usepackage{colortbl}
\usepackage{geometry}
\usepackage{times}
\usepackage{latexsym}
\usepackage{graphicx}
\usepackage{algorithm}
\usepackage{multirow}
\usepackage{amsfonts}
\usepackage{pgfplots}
\usepackage[T1]{fontenc}
\usepackage[utf8]{inputenc}

\usepackage{microtype}

\usepackage{inconsolata}

\usepackage{graphicx}
\usepackage{subcaption}
\usepackage{algorithm}
\usepackage{algpseudocode}
\usepackage{amsmath}
\usepackage{booktabs}

\usepackage[normalem]{ulem}
\useunder{\uline}{\ul}{}

\title{Memory-augmented Query Reconstruction for LLM-based Knowledge Graph Reasoning}

\author{Mufan Xu$^{\ast}$, Gewen Liang$^{\ast}$, Kehai Chen, Wei Wang, Xun Zhou\\ \textbf{Muyun Yang, Tiejun Zhao, Min Zhang}\\
  School of Computer Science and Technology, Harbin Institute of Technology, China \\
  \texttt{xmuffins0610@gmail.com, 24s051029@stu.hit.edu.cn},\\  \texttt{\{chenkehai,wangwei2019,zhouxun2023,yangmuyun,tjzhao,zhangmin2021\}@hit.edu.cn}}

\begin{document}
\maketitle

\begingroup\def\thefootnote{*}\footnotetext{Equal contribution.}\endgroup

\begin{abstract}
Large language models (LLMs) have achieved remarkable performance on knowledge graph question answering (KGQA) tasks by planning and interacting with knowledge graphs.
However, existing methods often confuse tool utilization with knowledge reasoning, harming readability of model outputs and giving rise to hallucinatory tool invocations, which hinder the advancement of KGQA.
%hinder
To address this issue, we propose \textbf{Mem}ory-augmented \textbf{Q}uery Reconstruction for LLM-based Knowledge Graph Reasoning (MemQ) to decouple LLM from tool invocation tasks using LLM-built query memory.
By establishing a memory module with explicit descriptions of query statements, the proposed MemQ facilitates the KGQA process with natural language reasoning and memory-augmented query reconstruction.
Meanwhile, we design an effective and readable reasoning to enhance the LLM's reasoning capability in KGQA.
%, significantly alleviating hallucinatory behaviors in existing methods.
% Experimental results show that MemQ not only strengthens the readability of the LLM-based knowledge graph reasoning process, but also enhances the stability of the KGQA process by ensuring precise tool invocation.
Experimental results that MemQ achieves state-of-the-art performance on widely used benchmarks WebQSP and CWQ.~\footnote{Our code and data will be released upon acceptance.}
%benchmark 写
\end{abstract}

\section{Introduction}
%第一段要写现有方法工具调用和推理一并进行，取得了一定的成就，中立的写
Large language models (LLMs) have demonstrated impressive reasoning capabilities in knowledge graph question answering (KGQA) task~\cite{yudecaf,huang2023towards,zhu2024benchmarking}. Using planning and interactive strategies, current LLM-based KGQA methods conduct the reasoning process on the knowledge graph based on the SPARQL tools and achieve remarkable performance across benchmarks~\cite{rog,ToG,xu2024generate}. 
Typically, part of these studies directly strengthens the reasoning ability of the LLM to plan tool-based paths and retrieve information from the knowledge graph~\cite{wang2023plan,rog}. The others employ LLMs to construct knowledge reasoning agents that execute the reasoning process on the knowledge graph through continuous tool-based decision-making based on environmental observations~\cite{gu2023don,jiang2024kg,xu2024llm}. These methods have achieved impressive results in the KGQA task.

\begin{figure}
    \centering
    \includegraphics[width=1\linewidth]{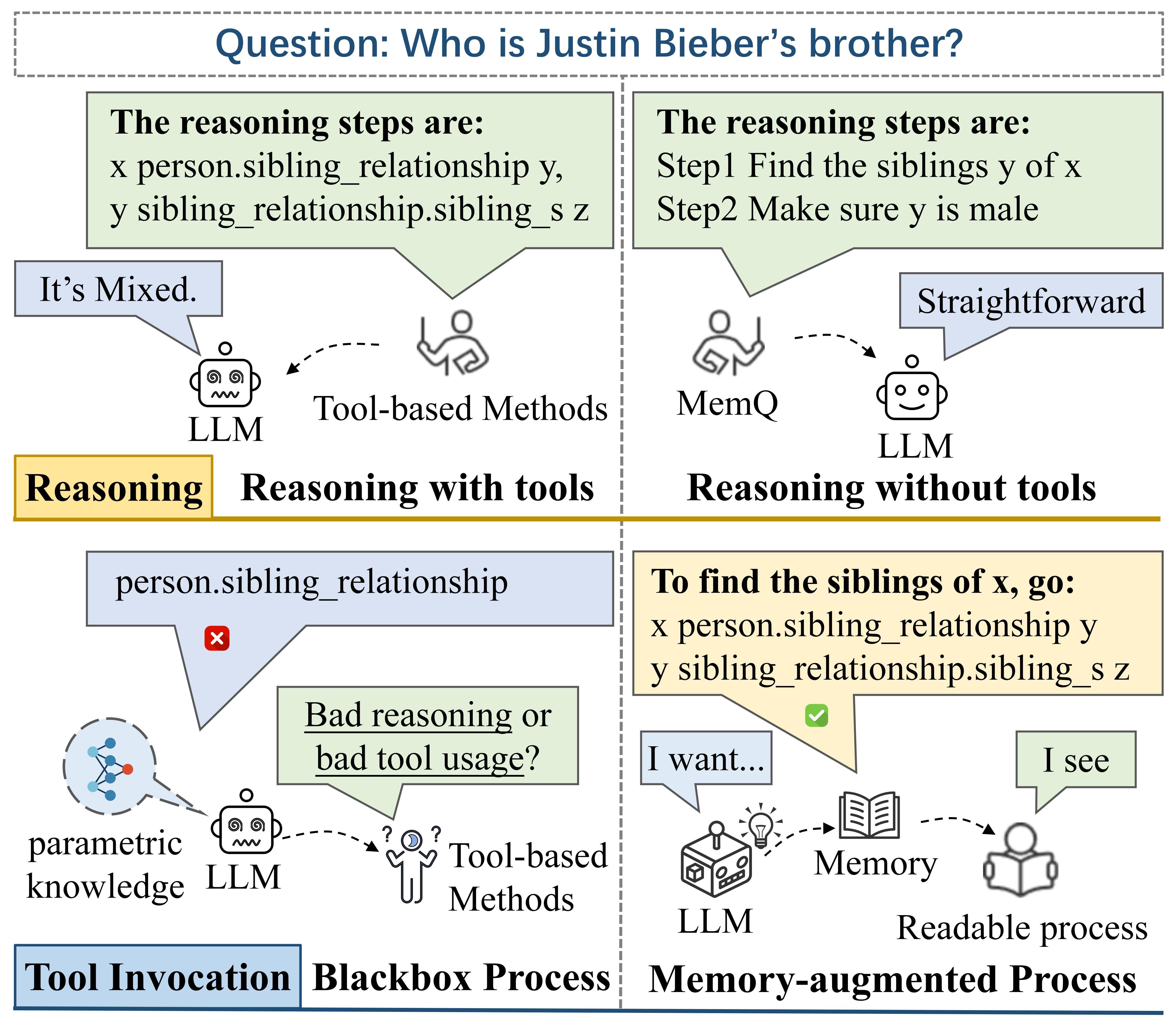}
    \caption{Comparing reasoning methods designed with knowledge graph query tools with proposed memory-augmented method MemQ.}
    \label{fig:motivation}
\end{figure}

However, the existing methods often confuse tool utilization with knowledge reasoning, harming readability and giving rise to hallucinatory tool invocations. As illustrated in Figure~\ref{fig:motivation}, when answering the question ``\textit{Who is Justin Bieber’s brother?}'' existing methods mixed tool invocation with knowledge reasoning tasks, which reduces the model's focus on the knowledge reasoning process (upper left). Furthermore, the mixed reasoning and tool invocation relies heavily on the LLM's parametric knowledge to utilizes the tool effectively, resulting in a black-box reasoning process with low interpretability (bottom left). Constructing a reasoning framework with tool invocation steps impairing readability and leading to erroneous tool invocations.

To address the issue, we propose \textbf{Mem}ory-augmented \textbf{Q}uery Reconstruction for LLM-based Knowledge Graph Reasoning (MemQ) to decouple LLM from the tool invocation task using an LLM-built query memory. To establish the query memory, we employ a rule-based strategy to decompose queries into statements, which are then described using the LLM’s capabilities, facilitating an independent reasoning process. We design an effective reasoning strategy based on natural language, enhancing readability and generating explicit reasoning steps. Based on the developed steps, MemQ retrieves memory based on semantic similarity and reconstructs the final query to interact with the knowledge graph. By establishing this query memory, the MemQ approach enables the model to disengage from tool invocation and focus on generating readable knowledge reasoning steps. Our main contributions are:
%这段冗余

\begin{itemize}
    \item We propose MemQ, a memory-augmented LLM-based KGQA reasoning framework to decouple reasoning from tool invocation task in the KGQA process.
%reduce the hallucinatory behavior brought by generation-based KGQA strategies.
%策略和幻觉
    \item The designed reasoning and memory construction strategies realize a readable LLM-based KGQA process, significantly alleviating the hallucinatory tool invocation issue.
    %By the discriminative reasoning framework, the proposed MemQ method not only enhances the capability of LLM to retrieve question-related subgraphs, but also alleviates the issue of ungrounded reasoning brought by the LLM generation process.
    \item The proposed MemQ achieved state-of-the-art performance on two widely used benchmarks WebQSP and CWQ.
\end{itemize}

\section{Related Works}

\textbf{Memory-augmented LLM Generation}.
Though large language models have demonstrated remarkable performance across tasks, they still struggle to achieve consistent performance on complex reasoning tasks~\cite{DBLP:conf/acl/ChenLCBXYZZ24,wang2024symbolic}. In this context, the approach of constructing an external knowledge base to record key information has been proposed and shown to be beneficial~\cite{hu2023chatdb,anokhin2024arigraph}. Researchers have proposed strategies to enhance LLM memory using external modules to support long-term dialogue history referencing~\cite{lee2024human,rezazadeh2isolated}. For tasks requiring extensive domain knowledge, methods for constructing memory banks either manually or using large language models have also been proven effective~\cite{cheng2024lift,panda2024holmes,edge2024local}.

\textbf{Knowledge Graph Question Answering}. Early KGQA approaches focused on using networks like key-value memory and graph neural networks to represent inference paths~\cite{miller-etal-2016-key,bai-etal-2022-graph,jiang2022unikgqa}, while other approaches teach models to build database queries such as SPARQL for direct answer retrieval ~\cite{gu-su-2022-arcaneqa,ye-etal-2022-rng}. With the rise of large language models (LLMs), methods utilize LLM's graph reasoning capability to enhance the reliability of reasoning process~\cite{zhong2024memorybank,zhang-etal-2024-question,zhu2024llms}. Certain approaches are developed to leverage scaled models to directly interact with Knowledge Graphs or for generating labels that assist smaller models in distilling reasoning abilities~\cite{ToG,xu2024generate}. Other efforts focus on constructing decision datasets based on annotated data to perform a supervised fine-tuning process, which enhance LLM's understanding of the knowledge reasoning process and their ability to interact with knowledge graphs~\cite{jiang2024kg}. Since LLM-generated outputs are generally susceptible to hallucinatory behavior, some research has shifted to employing discriminative strategies instead of generative ones to reduce unfounded reasoning processes~\cite{gu2023don,xu2024llm}.

\begin{figure*}
    \centering
    \includegraphics[width=0.85\linewidth]{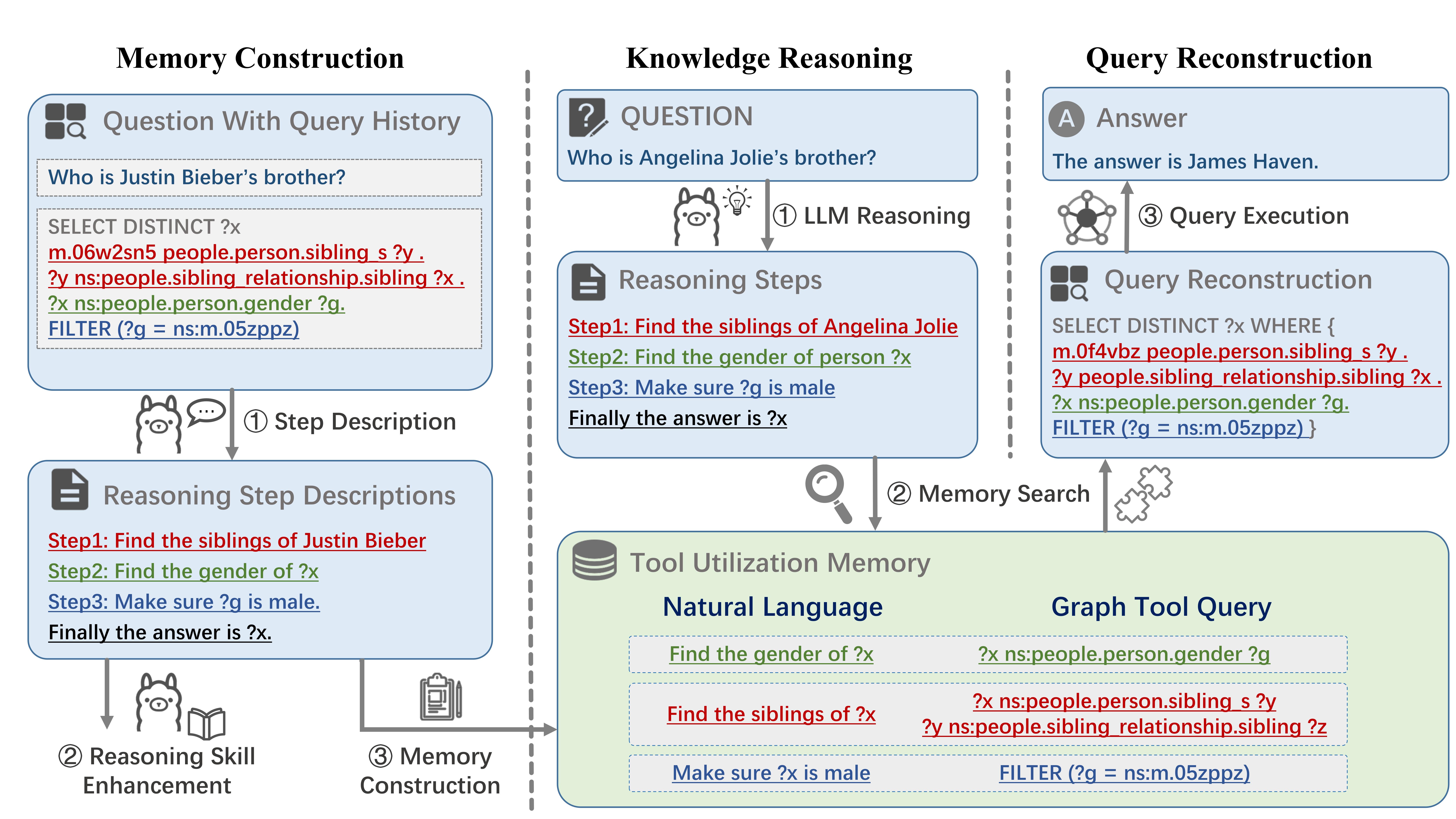}
    \caption{The overall framework of MemQ. During the memory construction stage, we describe the question with its query history using the LLMs to get the reasoning steps. In the inference stage, we reconstruct the query using the recalled query sentences based on the reasoning results.}
    \label{fig:total_framework}
\end{figure*}

However, the issue of confusing the tool invocation process with the knowledge reasoning process remains unresolved. The existing method often conducts reasoning based on SPARQL-formed edges like 'type.domain.property' or self-designed toolboxes, which diminishes the model's focus on the reasoning process and suffers from hallucinatory tool invocation behaviors. In this paper, we propose a memory-augmented KGQA reasoning method that effectively decouples the reasoning process from tool invocation.

\section{Framework with Memory Construction}

In this section, we introduce the framework of MemQ to decouple the reasoning process from tool invocation; the overall flow is illustrated in Figure~\ref{fig:total_framework}. We propose to facilitate the KGQA process using three tasks including memory construction, knowledge reasoning and query reconstruction. Before discussing the three tasks, we first illustrate the memory construction process.

\subsection{Memory Construction}

Given query history $H$ the contains question $q_i$ with its corresponding query $query_i$, the memory construction task asks the model to build a memory $M$ to represent the mapping function from natural language descriptions $n_i$ to query statements $s_i$:
\begin{equation}
\begin{aligned}
s_i=M(n_i),~s_i\in \text{query}_{i}.
\end{aligned}
\end{equation}

For example, if we have a question "\textit{what does x do?}" and its query ``\textit{select y where x people.person.career y.}'', we can directly save this pair of query and question into the memory $M$. It represents a mapping relationship from `\textit{one's job}' with query statement `\textit{x people.person.career y}'. It represents a mapping relationship from `x's career' with the query statement `\textit{select y where x people.person.career y}'.

\subsection{Knowledge Reasoning}

Given the question $Q$, the mentioned entities $E$, the knowledge reasoning task asks the model to develop an n-step reasoning plan $P$ to answer the question. Here we regulate $P$ with the rule that each reasoning step $p_i$ is limited to searching or examining only one entity. The n-step plan $P$ can be represented as a set of reasoning steps:
\begin{equation}
\begin{aligned}
P=\{p_i|i=1,2,...,n\}.
\end{aligned}
\end{equation}

For example, when answering the question ``\textit{Who is Justin Bieber's Brother?}'', the ideal reasoning step will start with the only known entity ``\textit{Justin Bieber}'' and search for the siblings of this person, and then we may figure out which one of the siblings is male to match with '\textit{brother}'. Note that we also need to record the retrieved new entities for potential use in subsequent steps, so we will always expect an assignment statement ``\textit{and assign it to <variable>.}'' in every search step. So we have $p_1=$``\textit{Find the siblings of Justin Bieber, assign it to x.}'', following by $p_2=$``\textit{Find the gender of person x, assign it to g.}''.

Reasoning steps that examine the answer or the value of a certain entity are often needed to meet the requirement of the question $Q$. In the previous example, we will have $p_3$=``\textit{Make sure g is male.}'' and $p_4$=``\textit{The answer is x.}'' to examine the value of \textit{g} and the position of the answer among known entities. Thus, a 4-step plan $P=\{p_1,p_2,p_3,p_4\}$ is given for the question.

\subsection{Query Reconstruction}

Given the developed reasoning plan $P$ and the query memory $M$, the query reconstruction task asks the model to first recall proper query statements $s_i$ using $M$ and then reconstruct the final query $Q_{f}$ corresponding to the question $Q$ using the set of collected statements:

\begin{equation}
\begin{aligned}
s_i=M(p_i),\\
Q_f=\text{Re-con}(S),\\
p_i\in P,~s_i\in S.
\end{aligned}
\end{equation}

Since the memory $M$ is constructed in a key-value form, we can directly recall the most similar memory using $M$ to reconstruct the new query. Referring to the example from the previous section, concerning the reasoning step ``\textit{find the gender of x}'', we expect the most similar memory to be recalled as ``\textit{x people.person.gender g.}''

\section{Approach}

After we propose the MemQ framework, we are able to design efficient strategies to facilitate memory construction, knowledge reasoning, and query reconstruction. 
Based on the tasks, we model the KGQA process as illustrated in Figure~\ref{fig:total_framework}.
% As illustrated in Figure~\ref{fig:total_framework}, the semantic query memory is constructed using step-level explanations of decomposed search queries. These explanations are then used to enhance the LLM's reasoning capability and construct the query memory bank. In the inference stage, the reasoning steps developed by the expert LLM is used to retrieve relevant query statements from the memory. The from Planning Expert LLM is processed by the Plan-to-SPARQL Module to produce a SPARQL query, which is executed on the Freebase Knowledge Graph to obtain the final answer.

\subsection{Memory Construction Strategy}

MemQ utilizes a rule-based strategy to decompose queries and then gather the description of each statement using the LLM. Based on the corresponding descriptions and statements, we establish a memory for the query statements to augment the query reconstruction process.

\noindent\textbf{Rule-based Decomposing.} Not every triplet in the knowledge graph conveys a readable meaning that can be described in natural language, which arises from the Compound Value Type (CVT) nodes that lack inherent semantic meanings. When splitting query statements, MemQ always uses non-CVT nodes as the starting or ending nodes, while regarding any encountered CVT nodes as intermediate nodes to ensure the semantic readability of individual statements. If no CVT node is encountered, the statement will contain only a single triplet.

\begin{figure}[h]
\begin{minipage}[h]{0.48\textwidth}
    \includegraphics[width=\textwidth]{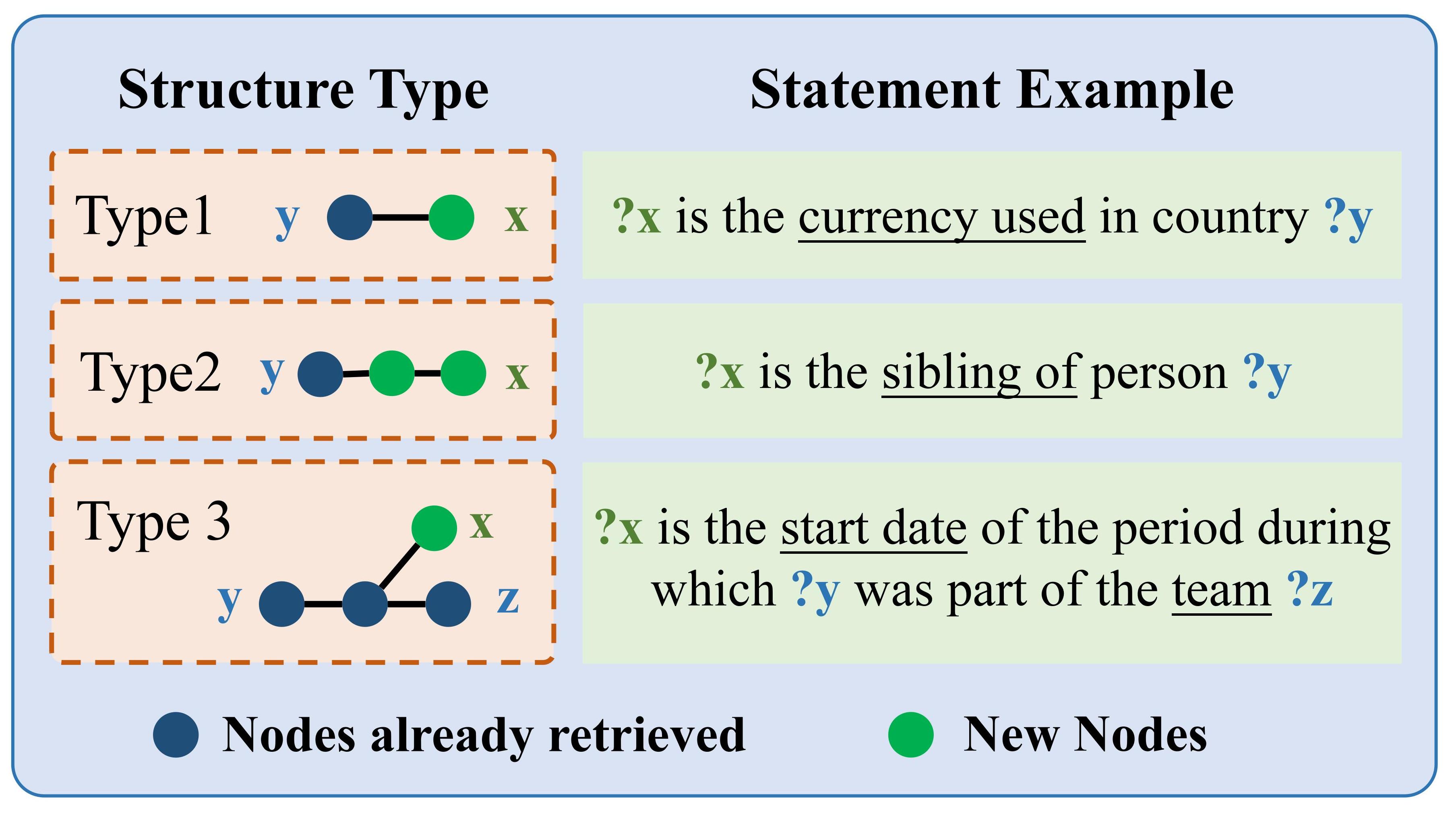}
    \caption{Here we present the illustration of the 3 distinct graph structures.}
    \label{fig:subgraph_structure}
\end{minipage}
\end{figure}

Using the strategy introduced, we can get query statements each of which stands for an operation with an atomic semantic message, such as "someone's hometown". As illustrated in Figure~\ref{fig:subgraph_structure}, the established memory contains statements with three distinct structures. By decomposing the queries, we obtained a total of 481 statements for type 1, 371 for type 2, and 142 for type 3.

\noindent\textbf{Description Collection.} For each statement, we use the LLM to provide a natural language description and store them in the query memory in pairs. We provide task instructions and examples in the context of conducting few-shot generation to ensure the quality of the description and prevent excessive differences between descriptions. We adopt GLM-4 as the description model to generate the descriptions. The prompt templates are shown in Appendix~\ref{appendix_prompt}. The memory construction process is actually a summarization and compression of historical search queries, providing readable hints to future query reconstruction process.

\subsection{LLM Reasoning in Natural Language}

As shown in Figure~\ref{fig:total_framework}, after obtaining the corresponding description of each query statement, MemQ uses those explanation-statement pairs to finetune the LLM to enhance its reasoning capabilities (bottom left). By adopting a memory-enhanced approach instead of using a model to directly generate query invocation content, MemQ only requires the LLM to focus on the reasoning process by generating reasoning steps based on the questions using natural language. The generated reasoning steps will be used for memory reconstruction process.

% Additionally, FILTER clauses in original SPARQL queries are translated into natural language using predefined patterns. Once all plans are generated, we fine-tune the Llama-3-instruct model to develop the Planning Expert.

\subsection{Query Reconstruction Strategy}

\begin{table*}[ht]
\centering
\scalebox{0.94}{
\begin{tabular}{llllll}
\toprule[1pt]
\multirow{2}{*}{\textbf{Method}} &  \multicolumn{2}{c}{\textbf{WebQSP}} & \multicolumn{2}{c}{\textbf{CWQ}} \\ \cline{2-5} 
 &Hits@1       &F1      &Hits@1       &F1 \\ \hline
Llama2-7b zero-shot (\citealp{touvron2023llama})* &0.403       &0.293      &0.297       &0.272      \\
Llama3-8b zero-shot (\citealp{dubey2024llama})* &0.303       &0.257      &0.305       &0.278      \\
Qwen2.5-7b zero-shot (\citealp{yang2024qwen2})* &0.284       &0.237      &0.259       &0.241      \\ \hline
KV-Mem (\citealp{miller-etal-2016-key}) &0.467       &0.345      &0.184       &0.157      \\
GraftNet (\citealp{sun-etal-2018-open}) &0.664       &0.604      &0.368       &0.327      \\
QGG (\citealp{lan-jiang-2020-query}) &0.730       &0.738      &0.369       &0.374      \\
NSM (\citealp{He_2021}) &0.687       &0.628      &0.476       &0.424      \\
SR+NSM (\citealp{zhang-etal-2022-subgraph}) &0.689       &0.641      &0.502       &0.471      \\
SR+NSM+E2E (\citealp{zhang-etal-2022-subgraph}) &0.695       &0.641      &0.493       &0.463      \\
DECAF (DPR+FiD-3B) (\citealp{yudecaf}) &0.821       &0.788      &-       &-      \\
UniKGQA (\citealp{jiang2022unikgqa}) &0.751       &0.702      &0.507       &0.480      \\
KD-CoT (\citealp{wang2023knowledgedrivencotexploringfaithful}) &0.686       &0.525      &0.557       &-      \\
ToG w/ChatGPT (\citealp{ToG}) &0.758       &-      &0.589       &-      \\
ToG w/GPT-4 (\citealp{ToG}) &0.826       &-      &0.676       &-      \\
KG-Agent (\citealp{jiang2024kg}) &0.833       &0.810      &0.722       &0.692      \\
RoG (Top-3 relation path) (\citealp{rog})* &0.795       &0.701      &0.567       &0.547      \\ \hline
MemQ (Ours)&\textbf{0.841}       &\textbf{0.858}      &\textbf{0.803}       &\textbf{0.830}      \\ \bottomrule[1pt]
\end{tabular}
}
\caption{The results of our method compared with previous approaches on WebQSP and CWQ. The asterisk * denotes the results we reproduced. Note that the Hits@1 result reported in the original RoG paper (WebQSP 0.857, CWQ 0.626) is not calculated in the right way, see the author's response \href{https://github.com/RManLuo/reasoning-on-graphs/issues/11}{here}.}
\label{table:main}
\end{table*}

During the query reconstruction process, MemQ iterates and alternates between the two sub-steps of memory recall and statements assembling according to the reasoning steps planned in the previous task, until the end of the reasoning steps is reached. As the query is reconstructed, it is executed to retrieve the final answer from the knowledge graph.

% During reasoning, we will first input the question and topic entity to the Planning Expert LLM, and output the plan to solve the problem. Then we parse the generated plan is transform into SPARQL in the Plan-to-SPARQL Module.

\noindent\textbf{Adaptive Memory Recall Strategy.} Given the developed reasoning steps, MemQ recalls relevant memory based on semantic similarity and employs rule-based methods to concatenate these statements to reconstruct a complete query. To measure the semantic similarity, we use Sentence-BERT to encode the reasoning steps and the explanations in the memory. Since the similarity scores of the top-N memory fragments can be nearly identical, MemQ adopts an adaptive recall strategy to retrieve the statements from the memory:
\begin{equation}
\begin{aligned}
N=
\begin{cases} 
1 & \text{if } \text{top-1 similarity} \geq \gamma_1, \\
k & \text{if } \text{top-1 similarity}<\gamma_1,
\end{cases} \\
k=\text{count}_{case}(\text{similarity}\geq\gamma_2).
\end{aligned}
\end{equation}

\noindent\textbf{Rule-based Reconstruct Strategy.} MemQ designs a rule-based reconstruction strategy where the most recently recalled sentence is appended to the end of the existing query. Note that we allow the LLM generate the names of unknown entities (e.g., ``\textit{person\_n}'') in the developed steps, the recalled statements will also be refilled using those names. 
% retrieve more than one memory  When the top-1 memory similarity is sufficiently high, selecting only the top-1 memory might not lead to errors. However, when the top-1 memory similarity is insufficient, this greedy recall strategy could result in the loss of critical query fragments. To address this, we have developed an adaptive recall strategy to mitigate the impact of the aforementioned issue. If the top semantic similarity exceeds a predefined threshold $\gamma_1$, the top-ranked query is selected. Otherwise, all statements with semantic similarity above a secondary threshold $\gamma_2$ are retained. For steps that impose constraints on the results, predefined rules are applied to transform them into FILTER clauses within the query.

\section{Experiment}

In this section, we first introduce the datasets and evaluation methods used by MemQ. After presenting the main experimental results, we will follow up with reports on several analytical experiments to examine the characteristics of the MemQ method compared to previous methods from various perspectives.

\subsection{Benchmarks and Baselines}

\textbf{Benchmarks.} To evaluate the knowledge graph question-answering capability of the proposed method, we choose two widely used benchmarks, WebQSP (\citealp{webqsp}) and CWQ (\citealp{cwq}).

\noindent\textbf{Metrics.} We choose commonly used metrics Hits@1 and F1 for the evaluation process following previous works. For the definitions of metrics, please refer to Appendix~\ref{appendix:metrics}.

\noindent\textbf{Baselines.} We select previous SOTA approaches with tool-based strategies as baselines, including RoG with LLM planning and chain-of-thought reasoning strategy~\cite{rog}, ToG with interactive strategy~\cite{ToG}. We also list representative methods and zero-shot performances of widely used LLMs for comparison. We also finetune the LLMs with SPARQL queries for ablation, see Section~\ref{sec:ablation}.

\noindent\textbf{Base Model.} To ensure fairness in comparison, we choose Llama2-7b~\cite{touvron2023llama} as the base model following RoG~\cite{rog}. In analytical experiments, we adopt a stronger model Llama3-8b to better evaluate the effectiveness of our framework.

\subsection{Main Result}

\begin{table*}[ht]
\centering
\scalebox{0.85}{
\renewcommand{\arraystretch}{1.1}
\setlength{\tabcolsep}{3mm}
\begin{tabular}{lcccccccccccccc}
\toprule[1pt]
\multicolumn{1}{l|}{\textbf{Total Hops}}   & \textbf{1}    & \textbf{2}    & \textbf{3}    & \textbf{4}    & \textbf{5}    & \textbf{6}    & \textbf{7}    & \textbf{avg}   \\ \hline
\rowcolor{gray!30}
\multicolumn{9}{l}{\textbf{Edge Hitting Rate $EHR$}} \\ \hline
\multicolumn{1}{l|}{RoG}                       & \textbf{0.853}& 0.644         & 0.390         & 0.276         & 0.249         & 0.230         & 0.283         & 0.377\\
\multicolumn{1}{l|}{MemQ}                    & 0.816         & \textbf{0.844}& \textbf{0.854}& \textbf{0.851}& \textbf{0.854}& \textbf{0.861}& \textbf{0.939}& \textbf{0.860} \\ \hline
\rowcolor{gray!30}
\multicolumn{9}{l}{\textbf{Graph Edit Distance with Golden Graph ${GoldGED}$}} \\ \hline
\multicolumn{1}{l|}{RoG}                       & 0.479         & 2.494         & 3.764         & 4.505         & 5.499         & 7.193         & 10.438        & 4.910\\
\multicolumn{1}{l|}{MemQ}                    & \textbf{0.158}& \textbf{0.465}& \textbf{0.909}& \textbf{1.364}& \textbf{1.611}& \textbf{2.531}& \textbf{2.250}& \textbf{1.327} \\ \bottomrule[1pt]
\end{tabular}
}
\caption{We evaluate the Edge Hitting Rate and Graph Edit Distance with the golden graph for both our method and RoG. The results indicate that the reconstructed graphs achieve significantly higher accuracy and structural alignment compared to those generated by RoG.}
\label{table:exp_distance}
\end{table*}

The performance of our MemQ framework on the WebQSP and CWQ datasets is presented in Table~\ref{table:main}. Our method achieves state-of-the-art results on both benchmarks, as demonstrated by significant improvements in Hits@1 and F1 metrics. The results show the efficiency of proposed framework to decouple reasoning from tool invocation. By adopting a memory-augmented strategy, MemQ provides a new way to enhance the LLM-based reasoning process. 
% The performance enhancement is attributed to the Planning Expert model's proficiency in generating precise natural language reasoning steps, complemented by the query reconstruction module's proficiency in accurately reconstructing queries. As shown in the result, this synergy effectively mitigates the tool invocation hallucination problem prevalent in existing KGQA methods. 

\subsection{Reasoning Capability Analysis}

% As shown in Figure~\ref{table:main}, our method achieves a significantly higher F1 score compared to other baselines, suggesting that the answers generated by our MemQ approach frequently encompass the majority of the content present in the ground truth answers. 
To investigate the improvements brought by our proposed reasoning framework, we conduct experiments to examine the discrepancies between the search graph of the reconstructed queries and that of the golden queries. We evaluate the quality of the developed subgraph from two aspects: 1) the structural accuracy and 2) the edge accuracy. Our analysis specifically targets these dimensions to identify the principal factors driving the observed performance improvements. 

The structural accuracy GoldGED is defined as the Graph Edit Distance between the reconstructed graph $G_{re}$ and the golden graph $G_{gd}$:
\begin{equation}
\begin{aligned}
\text{GoldGED}(G_{re}) = \min_{\pi \in \Pi(G_{re}, G_{gd})} \text{num}(\pi).
\end{aligned}
\end{equation}

The edge accuracy is quantified by the Edge Hitting Rate, which is computed using the hitting rate between edges in the golden graph $G_{gd}$ and the edges in the reconstructed graph $G_{re}$:
\begin{equation}
\begin{aligned}
\text{EHR}(G_{re}) = \frac{\text{num}(\{e| e\in G_{gd} \wedge e\in G_{re}\})}{\text{num}(\{e| e \in G_{gd}\})}.
\end{aligned}
\end{equation}

The reuslts is featured in Table~\ref{table:exp_distance}. Specifically, MemQ achieves a significantly lower GoldGED, indicating more accurate structural alignment with reference graphs, especially in complex multi-hop scenarios. Additionally, MemQ sustains a higher EHR, demonstrating robust edge accuracy even as the number of reasoning steps increases. Overall, these results emphasize MemQ's superior performance in producing accurate and structurally coherent graph-based reasoning across subgraphs.

\subsection{Ablation Study}
\label{sec:ablation}

To further analyze the effectiveness of the proposed framework, we conduct experiments to ablate the strategies in MemQ and observe the change in performance. We design two finetune-based baselines to ablate our strategies. 1) For the query reconstruction process, we directly finetune the model utilizing the statements and the descriptions recorded in the memory (denoted as -w/o QRM) to evaluate the effectiveness of our proposed query memory; 2) For the whole MemQ framework, we finetune the model using queries to simulate a straightforward tool-based reasoning process (denoted as -w/o PE, QRM) to evaluate the effectiveness of the MemQ framework. The results are shown in Table~\ref{table:exp_ablation}.

\begin{table}[h]
\centering
\scalebox{0.85}{
\setlength{\tabcolsep}{0.5mm}
\renewcommand{\arraystretch}{1.2}
\begin{tabular}{l|ccc|ccc}
\hline
\multirow{2}{*}{\textbf{Strategy}} & \multicolumn{3}{c|}{\textbf{WebQSP}} & \multicolumn{3}{c}{\textbf{CWQ}} \\ \cline{2-7} 
            & Hits@1    & F1        & EHR       & Hits@1    & F1        & EHR   \\ \hline
MemQ        & 0.857     & 0.872     & 0.858     & 0.817     & 0.845     & 0.886 \\ \hline
-w/o QRM    & 0.729     & 0.743     & 0.849     & 0.588     & 0.620     & 0.864 \\ \hline
-w/o PE,QRM & 0.733     & 0.731     & 0.739     & 0.556     & 0.570     & 0.806 \\ \hline
\end{tabular}
}
\caption{We conduct ablation studies to evaluate the impact of key components in our method by comparing it with two settings: 1) removing the Planning Expert (PE) and 2) removing both the Planning Expert (PE) and the Query Reconstruction Module (QRM).}
\label{table:exp_ablation}
\end{table}

\begin{figure*}[htbp]
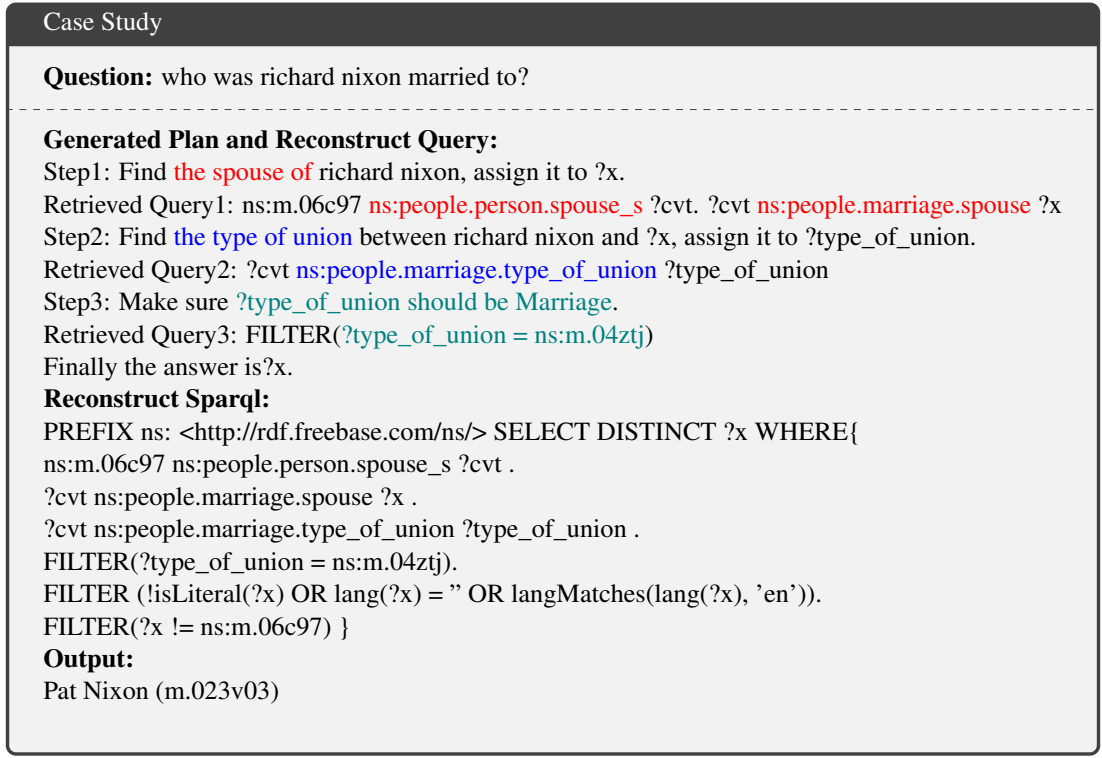

\centering
\scalebox{0.9}{
\begin{tcolorbox}[title = {Case Study}]
\textbf{Question:} who was richard nixon married to?
\tcblower
\textbf{Generated Plan and Reconstruct Query:} \\
Step1: Find \textcolor{red}{the spouse of} richard nixon, assign it to ?x. \\
Retrieved Query1: ns:m.06c97 \textcolor{red}{ns:people.person.spouse\_s} ?cvt. ?cvt \textcolor{red}{ns:people.marriage.spouse} ?x  \\
Step2: Find \textcolor{blue}{the type of union} between richard nixon and ?x, assign it to ?type\_of\_union. \\
Retrieved Query2: ?cvt \textcolor{blue}{ns:people.marriage.type\_of\_union} ?type\_of\_union \\
Step3: Make sure \textcolor{teal}{?type\_of\_union should be Marriage}. \\
Retrieved Query3: FILTER(\textcolor{teal}{?type\_of\_union = ns:m.04ztj})\\
Finally the answer is?x. \\
\textbf{Reconstruct Sparql:}\\
PREFIX ns: <http://rdf.freebase.com/ns/> 
SELECT DISTINCT ?x WHERE\{ \\
ns:m.06c97 ns:people.person.spouse\_s ?cvt . \\
?cvt ns:people.marriage.spouse ?x . \\
?cvt ns:people.marriage.type\_of\_union ?type\_of\_union . \\
FILTER(?type\_of\_union = ns:m.04ztj). \\
FILTER (!isLiteral(?x) OR lang(?x) = '' OR langMatches(lang(?x), 'en')). \\
FILTER(?x != ns:m.06c97) \} \\
 \textbf{Output:} \\
Pat Nixon (m.023v03)\\
\end{tcolorbox}
}
\caption{Case of MemQ, we retrieve memories based on the reasoning steps and reconstruct the final query.}
\label{fig:case_study}
\end{figure*}

According to the results, we can observe that: 1) Comparing MemQ with ``-w/o QRM'', the proposed memory-augmented strategy significantly improves the stability of tool utilization process compared with LLM-based finetuning strategy; 2) Comparing ``-w/o QRM'' with ``-w/o PE, QRM'', in the case of using a direct fine-tuning strategy, the method of direct fine-tuning that blends reasoning with tool invocation has lowered the overall F1 and EHR score. Furthermore, given that our method has also improved the overall Hits@1 and F1 scores compared to previous tool-based SOTA work, these results demonstrate the enhancement of the proposed decoupling strategy on the reasoning process of LLMs.

\subsection{Case Study}

To demonstrate the readability of the proposed MemQ method, we present a detailed case that highlight its capability to produce clear, logically consistent reasoning plans and accurate reconstruction queries. Figure~\ref{fig:case_study} provides an example question alongside the corresponding reasoning plan and reconstructed query, demonstrating its readability. 
Refer to Appendix~\ref{appendix:case} for more cases.

% Notably, the setting without the Query Reconstruction Module utilizes the same knowledge reasoning plans as MemQ but achieves lower performance. This demonstrates that fine-tuned LLM struggles to learn and generate tool invocation queries. The Edge Hit Rate (EHR) in Table~\ref{table:exp_ablation} for our method and the setting without QRM are nearly equivalent, both significantly surpassing the performance of the setting lacking both the PE and QRM. This demonstrates that the PE plays a pivotal role in enhancing the knowledge reasoning capabilities of the LLM, leading to substantial improvements in the knowledge reasoning process.

\subsection{Reasoning Hallucination Analysis}

To figure out the impact of our decoupled reasoning strategy on the hallucination issue, we manually check and evaluate the error cases of MemQ and the ``-w/o PE, QRM'' baseline proposed in the ablation study. To guarantee an objective evaluation, we established criteria to check with the cases: 1) \textbf{Correctness}: whether the main reasoning steps contain errors, 2) \textbf{Completeness}: whether the reasoning logic lacks necessary filtering conditions, and 3) \textbf{Redundancy}: whether the reasoning logic includes irrelevant or unnecessary filtering conditions. We randomly sample 100 cases from the test set to record the frequency of each of the errors. Note that one sample may contain multiple errors at a time.

\begin{table}[h]
\centering
\scalebox{0.83}{
\setlength{\tabcolsep}{0.5mm}
\renewcommand{\arraystretch}{1.2}
\begin{tabular}{l|ccc}
\hline
\textbf{Strategy}            & \textbf{Correctness}   & \textbf{Completeness}  & \textbf{Redundancy}    \\ \hline
MemQ        & \textbf{8}    & \textbf{16}   & 16            \\ \hline
-w/o PE,QRM & 39            & 41            & \textbf{9}    \\ \hline
\end{tabular}
}
\caption{We manually assess the reasoning plans based on Consistency, Completeness, and Redundancy, documenting the number of plans that exhibit errors in each of these categories.}
\label{table:exp_human}
\end{table}

As shown in Table~\ref{table:exp_human}, our method significantly reduces the number of Correctness and Completeness errors, while errors in Redundancy slightly increase. The increment in Redundancy errors stems from our retrieval strategy, justified by the presence of edges with similar semantic meanings in the Knowledge Graph (see Appendix~\ref{appendix:case} for details). The result indicates that our proposed decoupled reasoning strategy significantly reduces the errors brought by the confusing tasks, indicating an alleviation of the hallucinatory tool invocation issue.

% To validate our approach, we compare our method with two ablation baselines: 1) fine-tuning a separate LLM as a tool invocation expert to replace the symbolic query reconstruction module and 2) directly fine-tuning the LLM to generate SPARQL queries. We evaluate the Hits@1, F1 and EHR scores of these baselines on the WebQSP and CWQ datasets, with the results presented in Table~\ref{table:exp_ablation}. Given the inherent lack of interpretability in the knowledge reasoning process when LLMs generate SPARQL queries directly in the first baseline, we randomly sample 50 examples from each dataset and derive their corresponding reasoning plans based on our predefined rules. To ensure an objective evaluation, we manually assessed the quality of the reasoning plans based on three criteria: 1)\textbf{Correctness}: whether the main reasoning steps contain errors, 2)\textbf{Completeness}: whether the reasoning logic lacks necessary filtering conditions, and 3)\textbf{Redundancy}: whether the reasoning logic includes irrelevant or unnecessary filtering conditions.

\subsection{Data Efficiency Analysis}

\begin{figure}[h]
    \centering
    \begin{subfigure}[b]{0.47\columnwidth}
        \centering
        \begin{tikzpicture}
            \begin{axis}[
                width=4.6cm,  
                height=4.6cm, 
                legend pos=south east,
                ymajorgrids=true,
                grid style=dashed,
                xlabel={Proportion of Training Data (\%)},
                ylabel={Scores},
                xticklabels={$10$, $25$, $50$, $75$, $100$},
		      xtick={1,2,3,4,5},
                tick label style={font=\tiny},
                ylabel style={font=\scriptsize, yshift=-16pt},
                xlabel style={font=\scriptsize,yshift=8pt}, 
                ymin=0.65,
                ymax=0.9
            ]

            \addplot+[sharp plot, mark=square*,mark size=1.2pt,mark options={solid,mark color=red}, color=red] 
		coordinates
		{(1,0.718) (2,0.779) (3,0.822) (4,0.853) (5,0.857)};
        \addlegendentry{\scriptsize{Hits@1}}
            \addplot+[sharp plot, mark=triangle*,mark size=1.2pt,mark options={solid,mark color=blue}, color=blue] 
		coordinates
		{(1,0.740) (2,0.801) (3,0.840) (4,0.864) (5,0.872)};
        \addlegendentry{\scriptsize{F1}}
            \end{axis}
        \end{tikzpicture}
        \caption{WebQSP}
        \label{fig:sub-first1}
    \end{subfigure}
    \hspace{0.02\columnwidth}
    \begin{subfigure}[b]{0.47\columnwidth}
        \centering
        \begin{tikzpicture}
            \begin{axis}[
                width=4.6cm,  
                height=4.6cm,        
                legend pos=south east,
                ymajorgrids=true,
                grid style=dashed,
                xlabel={Proportion of Training Data (\%)},
                xticklabels={$10$, $25$, $50$, $75$, $100$},
		      xtick={1,2,3,4,5},
                tick label style={font=\tiny},
                xlabel style={font=\scriptsize,yshift=8pt},  
                ymin=0.65,
                ymax=0.9
            ]

            \addplot+[sharp plot, mark=square*,mark size=1.2pt,mark options={solid,mark color=red}, color=red] 
		coordinates
		{(1,0.683) (2,0.744) (3,0.787) (4,0.802) (5,0.803)};
        \addlegendentry{\scriptsize{Hits@1}}
            \addplot+[sharp plot, mark=triangle*,mark size=1.2pt,mark options={solid,mark color=blue}, color=blue] 
		coordinates
		{(1,0.732) (2,0.783) (3,0.818) (4,0.831) (5,0.845)};
        \addlegendentry{\scriptsize{F1}}

            \end{axis}
        \end{tikzpicture}
        \caption{CWQ}
        \label{fig:sub-first}
    \end{subfigure}
    \caption{We evaluate the Hits@1 and F1 scores of the LLaMA-3 Reasoning LLM across varying proportions of training data.}
    \label{fig:exp_portion}
\end{figure}
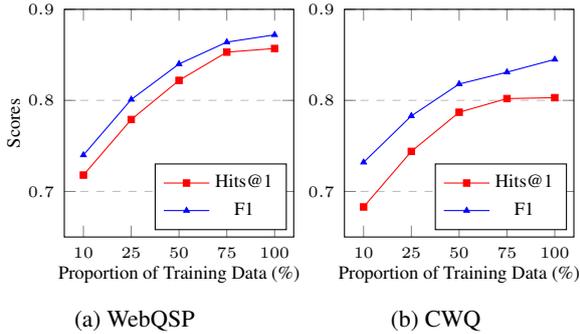
To assess the data efficiency of our MemQ method, we evaluate the performance of planning expert LLM trained with varying levels of training data availability. In this experiment, we randomly selected 10\%, 25\%, 50\%, 75\%, and 100\% of the step description data to fine-tune the LLaMA-3-8B-Instruct model. As illustrated in Figure~\ref{fig:exp_portion}, our method achieves an F1 score and Hits@1 of approximately 0.7 with only 10\% of the training data, significantly outperforming the zero-shot baseline across both datasets. Furthermore, performance improves steadily as the proportion of training data increases, indicating the method's ability to scale effectively with additional data. These results show that our method can effectively utilize limited data, highlighting its strong data efficiency, with consistently improved performance among different volume of training data.

\subsection{Model Universality Analysis}
\begin{table}[h]
\centering
\scalebox{0.90}{
\renewcommand{\arraystretch}{1.2}
\begin{tabular}{l|cc|cc}
\hline
\multirow{2}{*}{\textbf{Base Model}} & \multicolumn{2}{c|}{\textbf{WebQSP}} & \multicolumn{2}{c}{\textbf{CWQ}} \\ \cline{2-5} 
            & Hits@1    & F1        & Hits@1    & F1        \\ \hline
Vicuna-7b   & 0.828     & 0.846     & 0.796     & 0.826     \\ \hline
Llama2-7b   & 0.841     & 0.858     & 0.803     & 0.830     \\ \hline
Llama3-8b   & 0.858     & 0.872     & 0.818     & 0.845     \\ \hline
Qwen2.5-7b  & 0.828     & 0.850     & 0.793     & 0.818     \\ \hline
\end{tabular}
}
\caption{We fine-tuned four widely-used LLMs to assess method versatility, with all models demonstrating strong performance, confirming the approach's robustness across diverse architectures.}
\label{table:exp_base}
\end{table}

To demonstrate the robustness and versatility of our MemQ, we conduct fine-tuning experiments on four distinct, widely-used large language models (LLMs) serving as the Planning Expert to generate the reasoning steps. The results in Table~\ref{table:exp_base} demonstrate that all models achieved strong performance, indicating its adaptability to different LLM architectures and confirming its robustness as a model-agnostic solution for reasoning tasks.

\subsection{Error Analysis}

To conduct a detailed error analysis, we categorize errors into two distinct types: 1) \textbf{Main Path Error}, where the primary reasoning path is incorrect, and 2) \textbf{Filtering Error}, which includes cases of excessive or insufficient filtering. This classification allows for a systematic evaluation of the inaccuracies in the reasoning process. 

\begin{figure}[htb]
    \includegraphics[width=0.95\linewidth,height=0.3\textwidth]{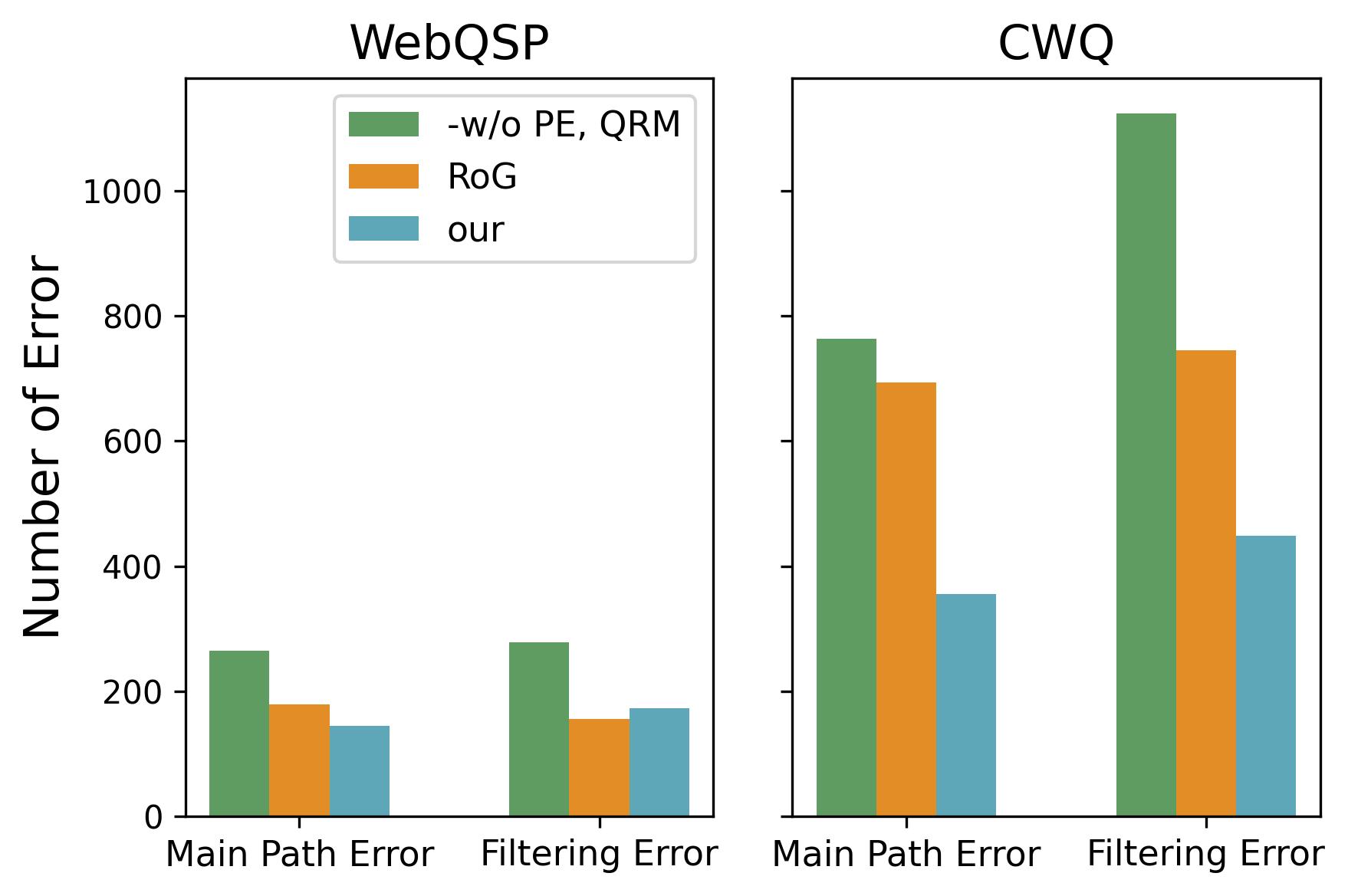}
    \caption{We compare our method with two baselines across two datasets, analyzing the number of errors categorized into two distinct types.}
    \label{fig:exp_error}
\end{figure}

As shown in Figure~\ref{fig:exp_error}, the Main Path Error of our method is significantly lower than the other two baselines in all datasets. In the CWQ dataset, our method achieves the lowest filtering error among all compared approaches. In the WebQSP dataset, our method achieves substantially lower filtering error compared to the setting without PE and QRM, though it is marginally higher than the RoG method. These results demonstrate the effectiveness of our method in reducing reasoning and filtering errors.

\section{Conclusion}

In this paper, we propose decoupling LLM from tool invocation tasks using an LLM-built query memory to alleviate hallucinatory tool invocation issues.
By facilitating the KGQA process using three tasks, we established a memory module to augment the query reconstruction process in the KGQA task. Based on the framework, we design an effective and readable reasoning strategy to enhance the LLM's reasoning capability, which also alleviates hallucinatory behaviors in existing methods. Experimental results show that our proposed memory-enhanced framework has achieved the state-of-the-art (SOTA) performance on two commonly used benchmarks.

\section*{Limitation}

Though our proposed MemQ framework has shown competitive KGQA performance and is proven to enhance the LLM’s reasoning capability, we identify several limitations that require further improvement. In the future, we will focus on the following directions to extend the current work:

\noindent1) Usage of Labeled data: Although our method effectively enhances LLM-based KGQA reasoning process and alleviates the hallucinatory tool invocations, we assume that we have the gold queries to construct the memory. However, it is noteworthy that the decomposing process of the query can be replaced by gathering all the relations and examples of the usage of relations from the Freebase itself. In the future, we will analyze the possibility of model the whole Freebase into a memory to get rid of the demand of gold queries.

\noindent2) Plug-and-play Capability: The proposed framework possesses good plug-and-play capability since the constructed memory is a portable module that can be adopted with other reasoning strategies and other tools. In the future, we will conduct experiments to showcase this kind of capability and testify our proposed memory-based framework under multi-tool or task transfer conditions.

\bibliography{custom}

\appendix

\section{Metrics}
\label{appendix:metrics}
In this section, we present the mathematical formulations and explanations for the metrics that were not fully elaborated in the main text.

\noindent\textbf{Hits@1.} Hits@1 quantifies the proportion of questions for which the top-ranked answer in the model's output is correct. Let $Answer$ represent the list of predicted answers, $Golden$ denote the list of ground truth answers, and $total\_num$ represent the total number of questions in the dataset. The formula is defined as follows: The formula of Hits@1 is defined as follows:
\begin{equation}
    Hits@1=\frac{count(Answer[0]\in Golden)}{total\_num}.
\end{equation}

\noindent\textbf{F1.} Following previous methods, we use the Macro-F1 scoring method, which calculates the F1 for each test sample and then averages those F1 scores among the samples.

\section{Prompt Template}
\label{appendix:prompt}

The used prompt templates are listed in the following tables. We designs 3 templates for the three types of queries shown in Table~\ref{table: Structure_1}, Table~\ref{table: Structure_2} and Table~\ref{table: Structure_3}. Besides, for the finetuning process to enhance the LLM's reasoning ability, we use the template in Table~\ref{table: finetune}.

\label{appendix_prompt}
\begin{table*}[htb]
\centering
\begin{tabular}{l}
\toprule[1pt]
\textbf{Prompt for Structure 1}   \\ \toprule[1pt]
Act as a SPARQL expert. \\I need you to explain the meaning and function of a specific part of a SPARQL query.\\
You job is answer the Question for me. ONLY OUTPUT THE ANSWER, NOTING ELSE!!\\

\#\#\# EXAMPLE1\\
Sparql:\\
?entity1 ns:location.country.currency\_used ?entity2 .\\
Question: How does ?entity2 related to ?entity1 ? \\Please answer the question with "?entity2 is [noun phrase]" .\\
Answer: ?entity2 is the currency used in the country ?entity1.\\

\#\#\# EXAMPLE2\\
Sparql:\\
?entity2 ns:location.country.currency\_used ?entity1 .\\
Question: How does ?entity2 related to ?entity1 ? \\Please answer the question with "?entity2 is [noun phrase]" .\\
Answer: ?entity2 is the country that use ?entity1 as currency.\\

\#\#\# EXAMPLE3\\
Sparql:\\
?entity2 ns:government.election\_campaign.candidate ?entity1 .\\
Question: How does ?entity2 related to ?entity1 ? \\Please answer the question with "?entity2 is [noun phrase]" .\\
Answer: ?entity2 is the election campaign which ?entity1 is the candidate.\\

\#\#\# EXAMPLE4\\
Sparql:\\
?entity1 ns:government.election\_campaign.candidate ?entity2 .\\
Question: How does ?entity2 related to ?entity1 ? \\Please answer the question with "?entity2 is [noun phrase]" .\\
Answer: ?entity2 is the candidate in the election campaign ?entity1.\\

\#\#\# EXAMPLE5\\
Sparql:\\
\{ ?entity2 ns:sports.sports\_championship\_event.runner\_up ?entity1 \} UNION \\\{ ?entity2 ns:sports.sports\_championship\_event.champion ?entity1 \}\\
Question: How does ?entity2 related to ?entity1 ? \\Please answer the question with "?entity2 is [noun phrase]" .\\
Answer: ?entity2 is either the runner-up or the champion of a sports championship event ?entity1.\\

\#\#\# EXAMPLE6\\
Sparql:\\
\{ ?entity1 ns:location.statistical\_region.places\_exported\_to ?tmp0 . \\?tmp0 ns:location.imports\_and\_exports.exported\_to ?entity2 \} UNION\\\{ ?entity1 ns:location.statistical\_region.places\_exported\_from ?tmp1 . \\?tmp1 ns:location.imports\_and\_exports.exported\_from ?entity2 \}\\
Question: How does ?entity2 related to ?entity1 ? \\Please answer the question with "?entity2 is [noun phrase]" .\\
Answer: ?entity2 is the place that is either exported to or exported from the statistical region ?entity1.\\

\#\#\# YOUR TURN\\
Sparql:\\
\{sparql\}\\
Question: How does ?entity2 related to ?entity1 ? \\Please answer the question with "?entity2 is [noun phrase]" .\\
Answer: \\
\bottomrule[1pt]
\end{tabular}
\caption{The prompt to get the explanation of Structure 1 graph}
\label{table: Structure_1}
\end{table*}

\begin{table*}[htb]
\centering
\begin{tabular}{l}
\toprule[1pt]
\textbf{Prompt for Structure 2}   \\ \toprule[1pt]
Act as a SPARQL expert. \\I need you to explain the meaning and function of a specific part of a SPARQL query.\\
You job is answer the Question for me. ONLY OUTPUT THE ANSWER, NOTING ELSE!!\\

\#\#\# EXAMPLE1\\
Sparql:\\
?cvt ns:government.government\_position\_held.office\_holder ?entity1 .\\
?entity2 ns:government.governmental\_body.members ?cvt . \\
Question: How does ?entity2 related to ?entity1 ? \\Please answer the question with "?entity2 is [noun phrase]" .\\
Answer: ?entity2 is the governmental body that is held by ?entity1.\\
Answer: ?entity2 is the governmental body that has an office holder ?entity1.\\

\#\#\# EXAMPLE2 
Sparql:\\
?entity1 ns:film.actor.film ?cvt .\\
?cvt ns:film.performance.character ?entity2 .\\
Question: How does ?entity2 related to ?entity1 ? \\Please answer the question with "?entity2 is [noun phrase]" .\\
Answer: ?entity2 is the character played by the actor ?entity1.\\

\#\#\# EXAMPLE3 \\
Sparql:\\
?cvt ns:music.group\_membership.member ?entity1 .\\
?entity2 ns:music.musical\_group.member ?cvt .\\
Question: How does ?entity2 related to ?entity1 ? \\Please answer the question with "?entity2 is [noun phrase]" .\\
Answer: ?entity2 is the musical group that has the member ?entity1.\\
Answer: ?entity2 is the group that includes the member ?entity1.\\

\#\#\# YOUR TURN\\
Sparql:\\
\{sparql\}\\
Question: How does ?entity2 related to ?entity1 ? \\Please answer the question with "?entity2 is [noun phrase]" .\\
Answer:  \\
\bottomrule[1pt]
\end{tabular}
\caption{The prompt to get the explanation of Structure 2 graph}
\label{table: Structure_2}
\end{table*}

\begin{table*}[htb]
\centering
\begin{tabular}{l}
\toprule[1pt]
\textbf{Prompt for Structure 3}   \\ \toprule[1pt]
Act as a SPARQL expert. \\I need you to explain the meaning and function of a specific part of a SPARQL query. \\
You job is complete the answer for me. ONLY OUTPUT THE ANSWER, NOTING ELSE!! \\
\#\#\# EXAMPLE1 \\
Sparql: \\
?cvt ns:sports.sports\_team\_coach\_tenure.position ?entity1 . \\
?cvt ns:sports.sports\_team\_coach\_tenure.coach ?entity2 . \\
?entity3 ns:sports.sports\_team.coaches ?cvt . \\
Question: How does ?entity3 related to ?entity1 and ?entity2 ? \\Please answer the question with "?entity3 is [noun phrase]" .  \\
Answer: ?entity3 is the sports team that has a coach ?entity2 who holds the position ?entity1 . \\
\#\#\# EXAMLPE2 \\
Sparql: \\
?entity1 ns:film.actor.film ?cvt . \\
?cvt ns:film.performance.character ?entity2 . \\
?cvt ns:film.performance.film ?entity3 . \\
Question: How does ?entity3 related to ?entity1 and ?entity2 ? \\Please answer the question with "?entity3 is [noun phrase]" . \\
Answer: ?entity3 is the film in which the actor ?entity1 performs the character ?entity2. \\
Answer: ?entity3 is the film in which ?entity1 acted as a character ?entity2. \\
\#\#\# EXAMLPE3 \\
Sparql: \\
?entity1 ns:sports.pro\_athlete.teams ?cvt . \\
?cvt ns:sports.sports\_team\_roster.team ?entity2 . \\
?cvt ns:sports.sports\_team\_roster.from ?entity3 \\
Question: How does ?entity3 related to ?entity1 and ?entity2 ? \\Please answer the question with "?entity3 is [noun phrase]" . \\
Answer: ?entity3 is the starting date when ?entity1 (the professional athlete) was part of the team ?entity2. \\
Answer: ?entity3 is the start date of the period during which ?entity1 was part of the team ?entity2. \\
\#\#\# YOUR TURN \\
Sparql: \\
\{sparql\} \\
Question: How does ?entity3 related to ?entity1 and ?entity2 ? \\Please answer the question with "?entity3 is [noun phrase]" . \\
Answer:  \\
\bottomrule[1pt]
\end{tabular}
\caption{The prompt to get the explanation of Structure 3 graph}
\label{table: Structure_3}
\end{table*}

\begin{table*}[htb]
\centering
\begin{tabular}{l}
\toprule[1pt]
\textbf{Prompt for Plan Expert}   \\ \toprule[1pt]
You are given a problem to solve step by step. Each step should begin with either \\
"Find", "Make sure" or "Rank". Finally, you need to output which one is the final \\
answer. \\
The steps should logically follow from one another, where each step builds on the \\
outcome of the previous steps. \\
Each step should be simple, clear, and directly related to achieving the overall goal. \\
Some topic entities you can use to start the plan are provided below.\\
Question:\\
\{question\}\\
Topic Entities:\\
\{topic\_entities\}\\
\bottomrule[1pt]
\end{tabular}
\caption{The prompt utilized for generating knowledge reasoning plans in the Planning Expert.}
\label{table: finetune}
\end{table*}

\section{More Cases}
\label{appendix:case}
Here, we present two additional cases generated by our method. As shown in Table~\ref{table:case1}, our method accurately constructs queries with "Order By" and "Limit" clauses in Step 5, demonstrating its ability to interpret the temporal meaning of "last time" in the question, which a nuance often overlooked by previous methods. 

In Table~\ref{table:case2}, our method retrieves multiple queries with similar semantic meanings. While this approach may introduce redundancy, we argue that it is justified given the nature of the Freebase Knowledge Graph, where edges with similar semantic meanings do exist and can be challenging even for humans to distinguish. Consequently, retrieving all such edges ensures comprehensive coverage of potentially relevant answers.

\begin{table*}[]
\centering
\begin{tabular}{l}
\toprule[1pt]
\textbf{WebQTest-994}   \\ \toprule[1pt]
\textbf{Question:} \\ 
When was the last time the Dallas Cowboys won the superbowl? \\
\textbf{Generated Plan and Reconstruct Query:} \\
Step1: Find \textcolor{red}{the championship won by the sports team} dallas cowboys, assign it to ?x.\\
Retrieved Query1: ?entity1 \textcolor{red}{ns:sports.sports\_team.championships} ?entity2\\
Step2: Find \textcolor{blue}{the end date of the event} ?x, assign it to ?sk0. \\
Retrieved Query2: ?entity1 \textcolor{blue}{ns:time.event.end\_date} ?entity2\\
Step3: Find \textcolor{teal}{the championship associated with the sports championship event} ?x, assign \\it to ?championship. \\
Retrieved Query3: ?entity1 \textcolor{teal}{ns:sports.sports\_championship\_event.championship} ?entity2\\
Step4: Make sure \textcolor{red}{?championship should be super bowl}.\\
Retrieved Query4: FILTER(\textcolor{red}{?championship = ns:m.06x5s})\\
Step5: \textcolor{blue}{Sort the result based on ?sk0 in descending order} and \textcolor{teal}{keep the first result.}\\
Retrieved Query5: \textcolor{blue}{ORDER BY DESC(?sk0)} \textcolor{teal}{LIMIT 1}\\
Finally the answer is ?x.\\
% \textbf{Retrieved Query of Each Step:}\\
% Step1: ?entity1 ns:people.person.education ?cvt . ?cvt ns:education.education.institution ?entity2 .\\
% Step2: ?entity1 ns:common.topic.notable\_types ?entity2 .\\
% Step3: None \\
\textbf{Reconstruct Sparql:}\\
PREFIX ns: <http://rdf.freebase.com/ns/> SELECT DISTINCT ?x WHERE\{ \\ 
ns:m.02896 ns:sports.sports\_team.championships ?x . \\
?x ns:time.event.end\_date ?sk0 . \\
?x ns:sports.sports\_championship\_event.championship ?championship . \\
FILTER(?championship = ns:m.06x5s). \\
FILTER (!isLiteral(?x) OR lang(?x) = '' OR langMatches(lang(?x), 'en')). \\
FILTER(?x != ns:m.02896) \} \\ 
ORDER BY DESC(?sk0) LIMIT 1 \\
\textbf{Output:} \\
Super Bowl XXX (m.076wz)\\

\bottomrule[1pt]
\end{tabular}
\caption{Case Study 1}
\label{table:case1}
\end{table*}

\begin{table*}[]
\centering
\begin{tabular}{l}
\toprule[1pt]
\multicolumn{1}{l}{\textbf{WebQTest-212\_3c65a82425ff4488504b53ba502d0eb7}} \\ \toprule[1pt]
\textbf{Question:} \\ 
What is the name of the state that has the elected official Jan Brewer and also has the \\ Colorado River flow through it? \\
\textbf{Generated Plan and Reconstruct Query:} \\
Step1: Find the location that \textcolor{red}{partially contains} *Colorado River*, assign it to?x. \\
Retrieved Query1: \{?entity1 \textcolor{red}{ns:location.location.partially\_containedby} ?entity2 \} UNION\\
\{?entity2 \textcolor{red}{ns:location.location.partially\_contains} ?entity1 \}UNION\\
\{?cvt \textcolor{red}{ns:location.partial\_containment\_relationship.partially\_contains} ?entity1 . \\
?entity2 \textcolor{red}{ns:location.location.partiallycontains} ?cvt \}UNION\\
\{?entity1 \textcolor{red}{location.location.partially\_contained\_by} ?cvt \\ 
?cvt \textcolor{red}{ns:location.partial\_containment\_relationship.partially\_contained\_by}  ?entity2\}\\
Step2: Find the \textcolor{blue}{official who appointed the governing official} ?x, assign it to ?appointed\_by. \\
Retrieved Query2: ?entity1 \textcolor{blue}{ns:government.governmental\_jurisdiction.governing\_officials} ?cvt . \\?cvt \textcolor{blue}{ns:government.government\_position\_held.appointed\_by} ?entity2 . \\
Step3: Make sure \textcolor{teal}{?appointed\_by should be Jan Brewer}. \\
Retrieved Query3: FILTER(\textcolor{teal}{?appointed\_by = ns:m.02pkb1c})\\
Finally the answer is ?x. \\
\textbf{Reconstruct Sparql:}\\
\{ns:m.018qjq ns:location.location.partially\_containedby ?x \} UNION  \\
\{?x ns:location.location.partially\_contains ns:m.018qjq \} UNION  \\
\{?cvt ns:location.partial\_containment\_relationship.partially\_contains ns:m.018qjq . \\ \hspace{1em}?x ns:location.location.partiallycontains ?cvt \} UNION \\
\{ns:m.018qjq ns:location.location.partially\_contained\_by ?cvt1 . \\ \hspace{1em}?cvt1 ns:location.partial\_containment\_relationship.partially\_contained\_by ?x \}. \\
?x ns:government.governmental\_jurisdiction.governing\_officials ?cvt2 . \\ \hspace{1em}?cvt2 ns:government.government\_position\_held.appointed\_by ?appointed\_by . \\
FILTER(?appointed\_by = ns:m.02pkb1c). \\
FILTER (!isLiteral(?x) OR lang(?x) = '' OR langMatches(lang(?x), 'en')). \\
FILTER(?x != ns:m.018qjq)  \} \\
\textbf{Output:} \\
Arizona (m.0vmt) \\
\bottomrule[1pt]
\end{tabular}
\caption{Case Study 2}
\label{table:case2}
\end{table*}

\end{document}